%% file: main.tex
\documentclass[lettersize,journal]{IEEEtran}

\usepackage{amsmath,amsfonts}
\usepackage{amssymb}
\providecommand{\mathbbm}[1]{\ensuremath{\mathbf{#1}}}
\usepackage{algorithmic}
\usepackage{array}
\usepackage{graphicx}
\usepackage{textcomp}
\usepackage{xcolor}
\usepackage{booktabs}
\usepackage{multirow}
\usepackage{pifont}
\usepackage{tabularray}
\UseTblrLibrary{booktabs}
\usepackage{url}
\usepackage{cite}
\usepackage[bookmarks=false]{hyperref}
\usepackage{stfloats}
\usepackage{balance}
\usepackage{listings}
\lstdefinestyle{prompt}{
  basicstyle=\scriptsize\ttfamily,
  backgroundcolor=\color{gray!8},
  frame=single,
  rulecolor=\color{gray!40},
  breaklines=true,
  columns=fullflexible,
  keepspaces=true,
  aboveskip=4pt,
  belowskip=4pt,
}
\usepackage{placeins}

\providecommand{\eg}{\textit{e.g.}}

\graphicspath{{figures/}{figures/figs/}}

\hypersetup{pdftitle={Scene2Sound: Auditory-Grounded Soundscape Generation for 3D Gaussian Worlds},pdfauthor={Masaki Yoshida, Ren Togo, Takahiro Ogawa, Miki Haseyama}}
\begin{document}

\title{Scene2Sound: Auditory-Grounded Soundscape Generation\\ for 3D Gaussian Worlds}

\author{Masaki~Yoshida, Ren~Togo, Takahiro~Ogawa, and Miki~Haseyama\\
Hokkaido University
}

\maketitle

\begin{abstract}
3D Gaussian Splatting (3DGS) turns captured or generated imagery into photorealistic 3D world simulations that users can freely explore, yet these worlds remain silent. Because existing audio generation methods condition on a single image or viewpoint, their sound is tied to that observation and cannot stay consistent while a listener moves. We introduce the task of generating a spatially consistent soundscape for a given 3DGS world through auditory grounding, identifying which objects in the world should emit sound and anchoring each to a persistent 3D position, and present Scene2Sound, a training-free framework built on this grounding. From the input world alone, our pipeline selects viewpoints that jointly cover the scene, identifies sound-emitting objects with a vision-language model, and associates the multi-view detections into 3D instances through Gaussian set matching, which measures the overlap between the Gaussian sets that render each detection. Each source then receives generated audio that a standard object-based audio engine spatializes in real time at arbitrary listener poses. We further propose two spatial-consistency metrics, one testing whether rendered audio responds consistently to listener motion and one testing whether the claimed sources are supported by views held out from their placement. On a curated set of generated 3DGS worlds and on 3DGS scenes generated from real-world $360^\circ$ captures, Scene2Sound preserves the audio quality of strong per-viewpoint baselines while remaining spatially consistent where per-viewpoint and single-panorama pipelines do not, and a user study confirms the perceptual benefit. Project page: \url{https://masaki-lmd.github.io/scene2sound/}.
\end{abstract}


\IEEEpeerreviewmaketitle

\input{sections/intro.tex}
\input{sections/related.tex}
\input{sections/figures/fig_framework.tex}
\input{sections/method.tex}
\input{sections/setup.tex}
\input{sections/figures/fig_qualitative.tex}
\input{sections/figures/fig_instance_gallery.tex}
\input{sections/results.tex}
\input{sections/discussion.tex}

\appendices

\section{Dataset Details}
\input{sections/supp/A1_DatasetDetails}

\section{Automatic Viewpoint Selection}
\input{sections/supp/A2_ViewpointSelection}

\section{Audio Source Placement Details}
\input{sections/supp/A4_AudioPlacement}

\section{VLM Prompt Design}
\input{sections/supp/A5_VLMPromptDesign}

\section{Spatial-Consistency Metric Details}
\input{sections/supp/A6_POAMetricValidation}

\section{Subjective Evaluation Details}
\input{sections/supp/A7_SubjectiveEvaluation}

\section{Qualitative Results: VLM Scene Understanding}
\input{sections/supp/A8_QualitativeResults}

\section{D-SAV360 Evaluation Protocol}
\input{sections/supp/A11_DSAV360Protocol}

\section{SonoWorld Re-implementation Details}
\input{sections/supp/A12_SonoWorldReimpl}

\section{Stage-wise Diagnostics}
\label{sec:supp_error_budget}
Table~\ref{tab:error_budget} details the per-stage diagnostics summarized in the Robustness Analyses of the main paper; the $J_{\min}$ sweep and co-located-instance stress test of Section~\ref{sec:supp_placement} probe the association stage directly.
\input{sections/tables_error_budget.tex}

\section{Runtime Analysis}
\input{sections/supp/A9_RuntimeAnalysis}

\balance
\makeatletter
\let\tmm@thebibliography\thebibliography
\renewcommand{\thebibliography}[1]{\tmm@thebibliography{#1}\scriptsize\setlength{\itemsep}{0.3pt}}
\makeatother
\bibliographystyle{IEEEtran}
\bibliography{refs}

\end{document}

%% file: sections/intro.tex
\section{Introduction}\label{sec:intro}

3D Gaussian Splatting (3DGS)~\cite{kerbl20233dgs} has become a powerful representation for photorealistic, real-time-renderable 3D scenes. It now underlies a broad class of navigable 3D world simulations: scenes reconstructed from captured imagery, as well as scenes generated from text or images by recent methods~\cite{zhou2024dreamscene360,yu2025wonderworld} and commercial systems~\cite{marble}, with growing work on scene understanding, editing, and physical simulation~\cite{wang2025ag2aussian,zhao2025physsplat}. Yet however obtained, such a world remains purely visual, carrying appearance and geometry but no sound.

\input{sections/figures/fig_teaser.tex}

Truly immersive 3D worlds require convincing \textit{soundscapes} (collections of sounds that characterize an environment)~\cite{schafer1993soundscape}, rendered as \textit{spatial audio} that adapts in volume and direction as a listener freely navigates~\cite{naef2002spatialized}.

Recent visual-to-audio~\cite{li2025sounding} and spatial-audio~\cite{Dagli2025SEE2SOUND,liu2025omniaudiogeneratingspatialaudio} methods synthesize high-quality and even multi-channel audio from images, but they operate per viewpoint without explicit 3D sound-source modeling. Systems that do attach audio to 3D scenes target transient impact sounds~\cite{li2025visual} or voice-driven co-creation~\cite{de2025sonora} (Sec.~\ref{sec:related}). The recent SonoWorld~\cite{sonoworld2026} outpaints a single photograph into a panorama and generates an audio-visual scene anchored to that panoramic viewpoint. We instead target sound for an arbitrary, given 3DGS world, whether generated or reconstructed, that a listener can traverse freely: such a world cannot be summarized by a single panoramic observation; sound-emitting objects must therefore be discovered from multiple viewpoints and integrated into persistent 3D sources. What remains missing is a unified approach that automatically generates spatially consistent soundscapes across an entire navigable 3D environment.

Two challenges follow. Such worlds often lack canonical reference camera positions, and it is not obvious how to observe an expansive scene with sufficient coverage. Moreover, the detections collected across viewpoints must be organized into persistent instances. When a street contains several cars, observations of the same car from different viewpoints must merge into one source, while different cars must remain separate sources; failure of this multi-view correspondence directly causes spatial inconsistency. We further require training-free operation on any pre-trained 3DGS. While recent 3DGS instance methods~\cite{ye2024gaussiangrouping} advance segmentation through per-Gaussian optimization, they require retraining and do not address \textit{which} objects should emit sound.

To this end, we present \textbf{Scene2Sound}, a training-free framework that generates soundscapes for a given 3DGS world through auditory grounding (Fig.~\ref{fig:teaser}). It determines which objects should emit sound and anchors each to a persistent 3D position, so that the generated soundscape remains spatially consistent under free navigation. Because real-time spatialization at arbitrary viewpoints requires each source to carry a persistent 3D location, we adopt an object-based audio representation~\cite{pike2016objectbased}, in which each source is specified by its content and 3D position and an audio engine spatializes the sources relative to the listener at interactive rates. This formulation centers on source identification and 3D placement; propagation effects such as reverberation and occlusion are not modeled in this work (Sec.~\ref{sec:discussion}).
Scene2Sound resolves the two challenges above through multi-viewpoint association. We automatically select viewpoints that jointly cover the scene and identify sound events from the collected views. Observing the scene from multiple viewpoints inevitably yields overlapping detections of the same object, and these overlapping observations share contributing Gaussians. We exploit this overlap through Gaussian set matching (GSM), which measures Jaccard similarity between the Gaussian sets that the rasterizer records as rendering each detected region, associating multi-view detections into consistent 3D instances without learned features or per-Gaussian parameters. Scene2Sound orchestrates VLMs~\cite{bai2025qwen25vltechnicalreport}, segmentation models~\cite{carion2025sam3}, and text-to-audio models~\cite{Evans2025StableAudio} within this pipeline.

In summary, our contributions are as follows:
\begin{itemize}
    \item We formulate soundscape generation for a given 3DGS world as an auditory grounding problem, identifying sound-emitting objects and anchoring each to a view-consistent 3D position, and present Scene2Sound, a training-free framework that realizes this on pre-trained 3DGS worlds without per-scene training.
    \item We introduce Gaussian set matching (GSM), which associates cross-view detections into 3D instances via Jaccard similarity over the rasterizer's tile--Gaussian sets, requiring no learned features, per-Gaussian parameters, or per-scene optimization. With automatic viewpoint selection, GSM forms the backbone of Scene2Sound, orchestrating vision-language, segmentation, and text-to-audio models.
    \item We propose a two-axis evaluation of spatial consistency, pairing Listener-Motion Consistency (LMC), which tests response to listener motion, with Cross-View Grounding Consistency (CGC), which tests support from held-out views. We release SoundscapePLY, a curated testbed, and report transfer to real-world scenes.
\end{itemize}

%% file: sections/figures/fig_teaser.tex
\begin{figure}[t]
    \centering
    \includegraphics[width=0.92\linewidth]{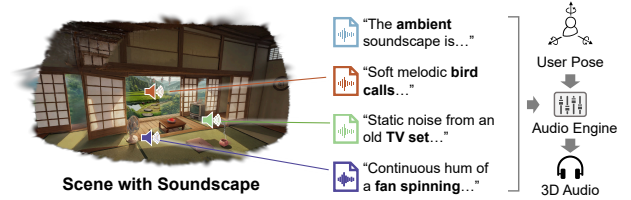}
    \caption{\textbf{Soundscape generation for 3DGS scenes.} Given a pre-trained 3DGS scene, Scene2Sound automatically infers \textit{what} sounds should be present and \textit{where} they originate, producing an object-based audio configuration. Combined with a user pose and an audio engine, this enables spatially consistent 3D audio that adapts as users navigate through the scene.}
    \label{fig:teaser}
\end{figure}

%% file: sections/related.tex
\section{Related Work}\label{sec:related}

\subsection{Audio Generation}
\label{sec:related_audio_generation}

Our task requires generating the audio content that composes a soundscape.
Recent generative audio models synthesize plausible waveforms from text or visual inputs at open-domain scale~\cite{liu2024audioldm2,Evans2025StableAudio}. Efficiency-oriented text-to-audio models such as AudioLCM~\cite{liu2024audiolcm}, FlashAudio~\cite{liu2024flashaudio}, and TangoFlux~\cite{hung2024tangoflux} reduce sampling cost, while Lumina-Next~\cite{zhuo2024lumina} is a general flow-based multimodal generation framework with an audio variant; video-to-audio models~\cite{NEURIPS2023_98c50f47,liu2025thinksound,liu2025prismaudio} such as MMAudio~\cite{cheng2025mmaudio} strengthen temporal and semantic alignment with visual content.
These approaches generate audio tied to the conditioning observation (a prompt, image, or video clip) rather than a reusable scene-level representation that supports re-rendering under viewpoint changes.

Two lines of work are closer to our setting.
Spatial audio generation methods directly produce immersive formats such as first-order ambisonics (FOA)~\cite{Heydari2025ImmerseDiffusion,liu2025omniaudiogeneratingspatialaudio} or 5.1-channel surround~\cite{Dagli2025SEE2SOUND}, but their outputs are anchored to the conditioning camera pose or equirectangular frame, giving viewpoint-locked waveforms.
SonoWorld~\cite{sonoworld2026} reconstructs a 3D audio-visual scene from a single image and synthesizes ambisonic audio for it, sharing our goal of spatially grounded sound; it regenerates the scene from one panoramic view, whereas we ground sound in a given 3DGS world and associate detections across views into object-based, re-renderable 3D instances, and we compare against a re-implementation of its pipeline in Sec.~\ref{sec:setup}.
Soundscape-oriented generation instead produces coherent environmental audio from object- or layer-level components, including object-aware generation conditioned on visual cues~\cite{li2025sounding}.

While these directions provide strong generators or composition paradigms, to our knowledge none associates multiple sound-emitting instances with persistent 3D anchors for consistent rendering under arbitrary listener motion; our pipeline supplies this authoring step.

\subsection{Spatial Audio Rendering in 3D Scenes}
\label{sec:related_audio_playback}

Audio rendering in 3D scenes involves two questions: what sound is produced where, and how it propagates to the listener.
Our work addresses the former, identifying and anchoring sound-emitting instances, and adopts object-based rendering for real-time playback.

Propagation- and field-based approaches model how sound travels through a scene, via impulse responses~\cite{wang2024hearing,ratnarajah2024avrir} or acoustic fields~\cite{luo2022NAF,lan2024acousticvolume}.
These methods capture room acoustics accurately but assume known sources or require multiple reference acoustic measurements of the target scene, a fundamentally different problem setting from ours.

Novel-view acoustic synthesis instead synthesizes audio from new listener viewpoints given recorded reference observations~\cite{chen2023novelview,liang2023AVNeRF}, including a material- and geometry-aware Gaussian representation~\cite{bhosale2024avgs} and a joint radiance--acoustic field informed by 3D scene structure~\cite{brunetto2025neraf}.
These methods reproduce view-dependent audio well but rely on captured audio or per-scene acoustic supervision (e.g., room impulse responses) and neural waveform synthesis, which is inapplicable when audio must be generated from scratch and played back efficiently under interactive viewpoint changes, as with generated or freshly reconstructed 3DGS worlds.

Object-based audio rendering instead represents a scene as a set of audio objects with associated metadata (source signals, 3D positions, rendering parameters), enabling real-time playback in audio engines~\cite{Geier2010Objectbased,Coleman2018AudioVisual}.
Our approach follows this paradigm, generating audio assets and estimating instance-level 3D anchors for a reusable scene-level representation.
Interactive authoring systems such as Sonora~\cite{de2025sonora} share this paradigm but rely on user-in-the-loop specification rather than fully automatic, scene-grounded anchoring.

\subsection{Object-Instance Retrieval in 3DGS Scenes}
\label{subsec:related_InstanceRetrieval}

Our framework requires separating sound-emitting object instances and estimating their 3D positions from a pre-trained 3DGS scene.
Most scene-segmentation methods~\cite{cen2025segment,Piekenbrinck2025OpenSplat3D,ye2024gaussiangrouping} obtain instance-level structure by augmenting Gaussians with learnable parameters optimized from multi-view 2D supervision, using per-Gaussian embeddings, identity encodings, or object-centric formulations, and thus require additional training beyond the original 3DGS reconstruction; object-aware Gaussian representations have also been explored for robotic manipulation~\cite{li2024objectaware}, under a different supervision setting.

Since our scenario assumes a reconstructed 3DGS and favors lightweight deployment, we avoid further training.
Training-free directions on 3DGS exist~\cite{jain2024gaussiancut,dai2025trainingfree} but are not tailored to audio-conditioned instance anchoring or to separating multiple instances under the same sound query.
Lifting by Gaussians (LBG)~\cite{chacko2025lifting} instead lifts 2D foundation-model signals into 3D, fusing masks and learned features to obtain instance-level predictions.
We avoid feature learning or fusion and instead perform training-free instance association by measuring Gaussian set overlap across views with the Jaccard index, directly grouping observations into sound-source instances atop the original 3DGS representation.
DCSEG~\cite{Wiedmann2025DCSEG} uses Jaccard-based matching to assign 2D open-vocabulary semantic labels to class-agnostic 3D masks; we instead use Gaussian-set overlap as the primary cue for cross-view instance association under a fixed sound query.

%% file: sections/figures/fig_framework.tex
\begin{figure*}[tp]
    \centering
    \includegraphics[width=0.78\textwidth]{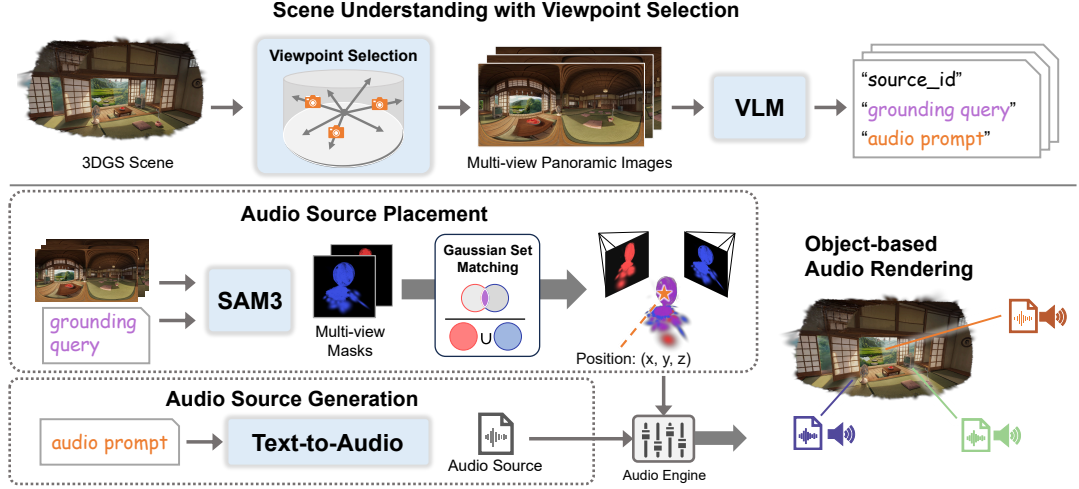}
    \caption{\textbf{Framework overview.} Our pipeline consists of three stages: (1)~\textit{Scene Understanding with Viewpoint Selection} renders multiple viewpoints from the 3DGS scene and infers sound events with their visual grounding using VLMs and segmentation models; (2)~\textit{Audio Source Placement} lifts 2D grounding regions into 3D via Gaussian set matching to produce spatially consistent instance associations; (3)~\textit{Audio Source Generation} synthesizes audio content for each placed source, yielding the object-based soundscape that an audio engine renders for the listener's pose.}
    \label{fig:framework}
\end{figure*}

%% file: sections/method.tex
\section{Method}\label{sec:method}

As shown in Fig.~\ref{fig:framework}, Scene2Sound addresses three challenges: inferring \textit{what} sounds should be present, determining \textit{where} each sound originates, and ensuring consistency across the navigable scene. We combine foundation models for perception and generation with algorithms that exploit 3DGS geometry to anchor sound sources in 3D.
We decompose soundscapes into \textit{ambient beds}, providing global background atmosphere without spatial anchoring (e.g., wind, distant traffic), and \textit{point sources}, spatially localized sounds tied to specific 3D positions (e.g., fountain, air conditioner). The decomposition follows from the rendering requirement: only sounds attributable to a 3D position can be spatialized relative to the listener, while atmosphere without a localizable origin is better rendered as a non-directional layer.

The pipeline has five modules: Automatic Viewpoint Selection (Sec.~\ref{sec:viewpoint}) selects observation cameras for scene coverage; Scene Understanding (Sec.~\ref{sec:understanding}) uses a VLM to identify sound events; Audio Source Generation (Sec.~\ref{sec:generation}) synthesizes audio from text prompts; Audio Source Placement (Sec.~\ref{sec:placement}) lifts 2D grounding into 3D positions via Gaussian set matching; and Audio Rendering (Sec.~\ref{sec:rendering}) integrates results into an audio engine for real-time playback.

\subsection{Problem Formulation}\label{sec:problem}

Given a 3DGS scene $\mathcal{G} = \{g_1, \ldots, g_{|\mathcal{G}|}\}$ of $|\mathcal{G}|$ Gaussian primitives, where each $g_i$ is parameterized by its center $\boldsymbol{\mu}_i \in \mathbb{R}^3$, opacity, and appearance attributes, we generate a soundscape
\begin{equation}
\mathcal{S} = (\mathcal{E}, a_{\mathrm{amb}}) = F(\mathcal{G}),
\end{equation}
where $\mathcal{E} = \{e_j\}$ is a set of positioned audio events, each $e_j = (a_j, \mathbf{p}_j)$ pairing an audio signal $a_j$ with a 3D position $\mathbf{p}_j \in \mathbb{R}^3$, and $a_{\mathrm{amb}}$ is a non-directional ambient sound. Unlike channel- or ambisonics-based outputs, which are tied to the viewpoint at which they were produced, this object-based representation can be re-rendered for any listener pose via standard audio engines, making it suitable for free navigation.

\subsection{Automatic Viewpoint Selection}\label{sec:viewpoint}

For generated 3DGS scenes, unlike reconstructed scenes, no reference camera positions exist. We automatically select $K$ camera positions $\{\mathbf{v}_1,\ldots,\mathbf{v}_K\}$ that jointly maximize \textit{visual quality} (for robust VLM predictions) and \textit{scene coverage} (for comprehensive audio event detection), using only static Gaussian parameters before rendering.

In typical navigable 3DGS scenes, Gaussians form surfaces surrounding an open interior space through which users move; we assume this hollow structure and design viewpoint selection accordingly. We compute a robust, outlier-filtered scene center $\mathbf{c}$, cast $N$ uniformly distributed rays from $\mathbf{c}$, and place one candidate camera per ray toward the scene shell. For each candidate $\mathbf{v}_i$, we identify contributing Gaussians $\mathcal{G}^{\mathrm{vis}}_i \subseteq \mathcal{G}$ via differentiable-rendering visibility and evaluate visual quality $q_i$ with CLIP-IQA~\cite{wang2023clipiqa} (implementation details in the supplementary material). Let $\mathcal{K} \subseteq \{1, \ldots, N\}$ index the selected viewpoints; we greedily select $K$ viewpoints maximizing quality-weighted Gaussian coverage:
\begin{equation}
\mathrm{Coverage}(\mathcal{K}) = \frac{1}{|\mathcal{G}|} \sum_{j=1}^{|\mathcal{G}|} \max_{i \in \mathcal{K}} \left[ q_i \cdot w_d(d_{ij}) \cdot \mathbbm{1}_{j \in \mathcal{G}^{\mathrm{vis}}_i} \right],
\end{equation}
where $d_{ij} = \|\mathbf{v}_i - \boldsymbol{\mu}_j\|$ is the distance from camera $i$ to Gaussian $j$, and $w_d(\cdot)$ is a distance weight decaying with distance (inverse-distance weighting; supplementary material). We use $N{=}20$ and $K{=}5$. At each selected viewpoint, we render equirectangular panoramic images for downstream scene understanding.

\subsection{Scene Understanding}\label{sec:understanding}

Once viewpoints are selected, we render panoramic images from the 3DGS scene and analyze them with a Vision-Language Model (VLM), Qwen2.5-VL-32B~\cite{bai2025qwen25vltechnicalreport}, to infer sound events, identifying both what sounds should be present and where they originate for subsequent audio generation and spatial grounding.

We design a structured prompt instructing the VLM to distinguish two event types. Point sources are spatially localized sounds with identifiable visual sources (e.g., fountain, car engine) that require precise 3D placement. Ambient beds are global soundscape layers without specific spatial locations (e.g., wind, distant traffic) that contribute to overall atmosphere but do not require spatial grounding.

For each point source, the VLM assigns a \textsf{source\_id} shared by observations of the same semantic sound event across viewpoints; distinct physical instances within an event are resolved later by Gaussian set matching (Sec.~\ref{sec:placement}). It also generates three fields: (1) a \textsf{grounding query} as a simple noun phrase for the segmentation model, (2) an \textsf{audio prompt} providing a detailed description for text-to-audio synthesis, and (3) an \textsf{estimated\_loudness} (quiet, moderate, or loud) for audio rendering. For ambient beds, we specify only the audio prompt.
The VLM jointly analyzes all selected viewpoints and outputs a unified scene-level event collection; the prompt schema and design rules are in the supplementary material.

\subsection{Audio Source Generation}\label{sec:generation}

Given the VLM's audio prompts, we synthesize waveforms using Stable Audio Open~\cite{Evans2025StableAudio}, a latent diffusion-based text-to-audio model trained on large-scale audio data, generating 30-second, $44.1$\,kHz waveforms for all sources; point-source audio is converted to mono for 3D spatialization, while ambient-bed audio remains stereo for an immersive atmosphere.

\subsection{Audio Source Placement}\label{sec:placement}

We next lift 2D visual observations into consistent 3D sound source positions that remain stable across viewpoints.

\noindent\textbf{Visual Grounding via Segmentation.}
For each sound event, we perform visual grounding with a text-based segmentation model, SAM3~\cite{carion2025sam3}: given the event's grounding query, it outputs pixel-level masks with confidence scores, each processed individually as a separate observation. This produces a set of 2D masks $\{M_{k,m}\}$, where $k$ indexes cameras and $m$ indexes masks within each view.
We delegate localization to a dedicated segmentation model rather than the VLM's own bounding boxes, since lifting to Gaussians requires pixel-accurate masks (a coarse box would sweep in Gaussians from surrounding geometry), and box predictions are unreliable on equirectangular panoramas under strong distortion. When a grounding query matches several same-class instances, each mask becomes a separate observation; the ambiguity is resolved downstream, where Gaussian set matching merges only observations whose Gaussian sets overlap.

\noindent\textbf{Meta-based Gaussian Identification.}
To lift 2D masks into 3D, we use tile-based rendering metadata from gsplat~\cite{ye2025gsplat}: the rasterizer partitions the image plane into fixed-size screen-space tiles and records which Gaussians contribute to each tile during rendering. Since panoramic rendering uses oriented cubemaps aligned with each camera's forward direction, we query tile-Gaussian mappings under the same orientation, giving consistent correspondence between masks and Gaussian sets. This yields the set of Gaussians $\mathcal{G}_{k,m} \subseteq \mathcal{G}$ contributing to mask $M_{k,m}$'s pixels. Because tiles are coarser than mask boundaries, the lifted set can include background primitives (e.g., walls behind the target); a one-sided depth-consistency gate, relative to the median rendered depth inside the mask, removes them (supplementary material).

\noindent\textbf{Instance Association via Gaussian Set Matching (GSM).}
A key challenge is determining whether observations from different viewpoints refer to the same physical sound source or to distinct instances; we exploit the fact that observations of the same object share overlapping sets of Gaussians in the 3DGS representation.
This is also where the 3DGS representation supplies information that multi-view images plus per-view depth cannot: its shared, persistent set of scene primitives acts as a common index across views; correspondence can therefore be tested by set overlap directly, without feature matching, learned embeddings, or geometric reasoning about viewpoint changes.
We first group observations by their \textsf{source\_id} assigned in the scene understanding step (Section~\ref{sec:understanding}). Within each group, we compute the Jaccard similarity between Gaussian sets from different cameras:
\begin{equation}
J(\mathcal{G}_{k,m},\; \mathcal{G}_{k',m'}) = \frac{|\mathcal{G}_{k,m} \cap \mathcal{G}_{k',m'}|}{|\mathcal{G}_{k,m} \cup \mathcal{G}_{k',m'}|}.
\end{equation}
Observations from the same camera are treated as distinct instances by definition, since the segmentation model already separates them spatially within each view. Observations exceeding a Jaccard threshold $J_{\min} = 0.15$ across views are merged using Union-Find, yielding merged instance Gaussian sets
\begin{equation}
\mathcal{G}'_j = \bigcup_{(k,m) \in \mathcal{C}_j} \mathcal{G}_{k,m},
\end{equation}
where $\mathcal{C}_j$ is the set of observations clustered into instance~$j$.

\noindent\textbf{Position Estimation.}
For each instance, we compute the 3D position as a weighted centroid of the merged Gaussian set:
\begin{equation}
\mathbf{p}_j = \frac{\sum_{i \in \mathcal{G}'_j} w_i \cdot \boldsymbol{\mu}_i}{\sum_{i \in \mathcal{G}'_j} w_i},
\end{equation}
where $\boldsymbol{\mu}_i \in \mathbb{R}^3$ is the center of Gaussian $g_i$ and $w_i > 0$ is the number of rendering tiles in which $g_i$ appears, automatically recorded by the rasterizer, scaled by the confidence of the segmentation mask that contributed $g_i$.

Instances sharing the same \textsf{source\_id} reuse the audio signal $a_j$ generated in Section~\ref{sec:generation}, completing the positioned audio events $\mathcal{E} = \{e_j = (a_j, \mathbf{p}_j)\}$ defined in the problem formulation.

\subsection{Audio Rendering}\label{sec:rendering}

Generated audio signals and their 3D positions pass to an audio engine for real-time playback; as users navigate, the engine adjusts volume and spatialization from listener position and orientation. Base volume follows the VLM's \textsf{estimated\_loudness}; instances sharing a source label are energy-normalized by $1/\sqrt{n}$, with $n$ the number of instances sharing the label, so that repeated objects contribute the label's energy once rather than $n$ times, and the ambient bed plays at a fixed gain below the loudest sources. We implement a lightweight object-based renderer following the standard web-audio spatialization node (\textsf{PannerNode}). Quantitative evaluations use power-law distance attenuation with azimuth-based stereo panning; the interactive demo and subjective-study stimuli play positioned sources through the browser-native \textsf{PannerNode} with head-related transfer function (HRTF) filtering for headphone playback. The representation is compatible with any audio engine supporting 3D spatialization.

%% file: sections/setup.tex
\section{Experimental Setup}\label{sec:setup}

We assemble \textbf{SoundscapePLY}, a target-domain evaluation testbed of $24$ 3DGS scenes generated with Marble~\cite{marble}, curated to cover diversity in soundscape-related content (urban, indoor, and natural environments) and visual style; its generated nature lets us control this diversity, matching our generative-world setting. Each scene consists of a scene description (text prompt), a concept image used for generation, and the resulting 3DGS (PLY format); the paired description and image also serve as scene-level queries for semantic evaluation, which 3DGS assets collected from the web would not provide. Further details are in the supplementary material.
To assess generalization beyond generated content, we additionally evaluate on real-world 3DGS scenes reconstructed from $360^\circ$ imagery (D-SAV360), reported as a transfer/robustness check in Sec.~\ref{sec:results}; the protocol is detailed in the supplementary material.
The full pipeline takes $222$\,s per scene on a single NVIDIA RTX A6000 (per-stage breakdown in the supplementary material).

\noindent\textbf{Evaluation Metrics.}
We evaluate Scene2Sound from four perspectives: audio quality, semantic alignment, binaural cues, and spatial consistency.

\noindent\textbf{Audio Quality.}
We report Fréchet Audio Distance (FAD)~\cite{kilgour2019fad} with PANNs embeddings~\cite{kong2020panns,tailleur2024fadhuman}. For robustness, we evaluate against two reference sets: DCASE 2024 Task 7~\cite{lagrange2025dcase} (FAD\textsuperscript{D}), a sound scene synthesis challenge whose domain closely matches our task, and Clotho~\cite{drossos2020clotho} (FAD\textsuperscript{C}), a standard reference in text-to-audio generation. All methods use $24$ samples (one per scene); FAD from so few samples is noisy, and we therefore read it comparatively across methods under the identical protocol.

\noindent\textbf{Semantic Alignment.}
ImageBind~\cite{girdhar2023imagebind} (IB) measures image--audio cosine similarity; CLAP~\cite{wu2023clap} measures text--audio cosine similarity using the scene prompt. Methods that place sources in 3D are scored on the mixture rendered at each evaluation viewpoint, averaged over viewpoints; single-waveform baselines are scored on their generated waveform.

\noindent\textbf{Binaural Cues.}
We report two descriptive statistics of the rendered binaural audio: interaural decorrelation, computed as $1{-}\mathrm{IACC}$ with IACC the windowed interaural cross-correlation coefficient, and the mean absolute Interaural Level Difference (ILD, dB). Both indicate the presence of binaural cues; neither is a quality score with a preferred direction (Sec.~\ref{sec:results}).

\noindent\textbf{Spatial Consistency.}
Spatial consistency under listener navigation is central to our task and is not captured by established metrics. We introduce and validate our metrics for this property in Sec.~\ref{sec:poa_validation}.

\noindent\textbf{Baselines.}
We compare against eleven methods in two groups.
\textit{Non-spatial:} For text-to-audio: AudioLDM2~\cite{liu2024audioldm2}, Tango2~\cite{majumder2024tango2}, MMAudio\textsuperscript{T}~\cite{cheng2025mmaudio}, and Stable Audio Open (SAO)~\cite{Evans2025StableAudio}, the T2A model used in our pipeline. For vision-to-audio: MMAudio\textsuperscript{V}~\cite{cheng2025mmaudio}, Seeing\&Hearing~\cite{xing2024seeinghearing}, and Im2Wav~\cite{sheffer2023im2wav}. These methods generate audio directly from scene descriptions or images without 3D sound-source modeling; they serve as quality and semantic references for our object-based representation.
\textit{Spatial audio:} ViSAGe~\cite{kim2025visage} produces first-order ambisonics~(FOA) from video, See2Sound~\cite{Dagli2025SEE2SOUND} generates 5.1-channel surround via object-level composition from a single image, OmniAudio~\cite{liu2025omniaudiogeneratingspatialaudio} synthesizes FOA from panoramic imagery, and the recent SonoWorld~\cite{sonoworld2026} generates a 3D audio-visual scene anchored to a single panorama.
SonoWorld constructs its scene from a single panoramic viewpoint, whereas ours takes the 3DGS world itself as input; we re-implement its pipeline on our scenes using the scene-center view of our viewpoint selection as its panorama ($K{=}1$), keeping its scene-understanding prompts, depth-weighted mask lifting, and MMAudio generator. The remaining components (VLM, segmenter, binaural engine, listener positions) match ours and all other methods, reducing differences due to interchangeable modules; implementation details are in the supplementary material.
All methods are evaluated at the same set of viewpoints; vision-conditioned baselines receive the corresponding panoramic renderings, while text-to-audio baselines receive the scene text prompt. The published per-viewpoint baselines generate audio independently at each viewpoint without 3D sound-source modeling; the consequence of this for spatial-consistency evaluation is analyzed in Sec.~\ref{sec:poa_validation}.

%% file: sections/figures/fig_qualitative.tex
\begin{figure}[t]
    \centering
    \includegraphics[width=1.0\linewidth]{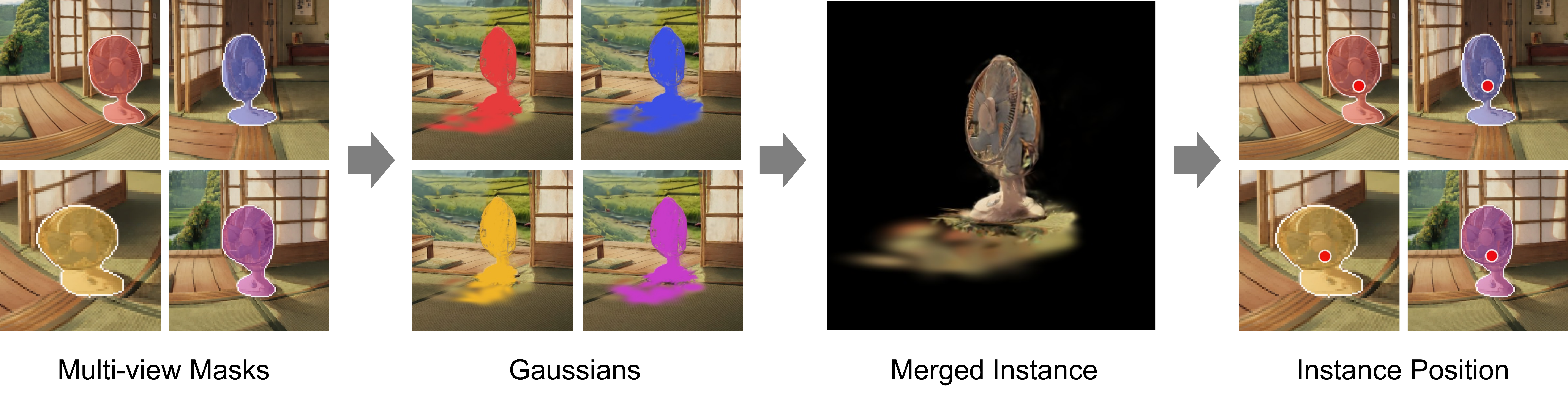}
    \caption{Instance association for a fan observed from four viewpoints. Per-view segmentation masks (left) are lifted to their corresponding Gaussian sets (middle-left), which are then merged into a single instance via Gaussian set matching (center). The resulting 3D instance position, projected back onto each view (right), consistently falls within the corresponding mask, illustrating how set overlap associates observations of one object across views.}
    \label{fig:qualitative}
\end{figure}

%% file: sections/figures/fig_instance_gallery.tex
\begin{figure}[t]
    \centering
    \includegraphics[width=\linewidth]{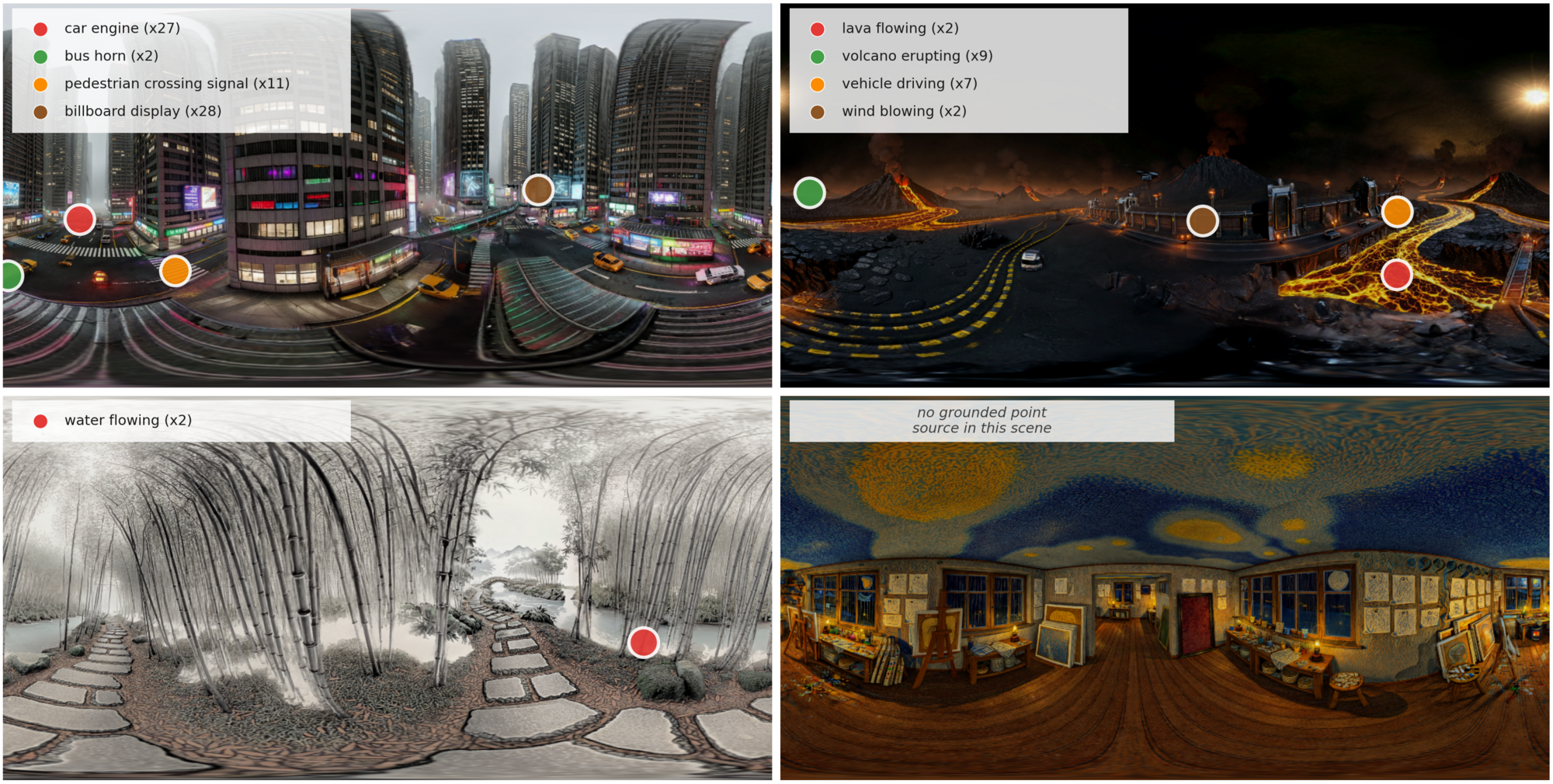}
    \caption{Instance association results across four representative scenes: projected 3D source positions on panoramic renderings, one representative marker per source label, with per-label instance counts in the legends. Bottom right: a failure case in which no sound-emitting object is grounded, so only the ambient bed is generated.}
    \label{fig:instance_gallery}
\end{figure}

%% file: sections/results.tex
\section{Results}\label{sec:results}

\subsection{Main Results}

\input{sections/tables_main_results.tex}

Table~\ref{tab:main_results} presents quantitative comparisons on SoundscapePLY. Bold marks the best among spatial audio methods, our primary comparison; shading ranks all methods, including the non-spatial quality references. Non-spatial methods are included as quality references, since they generate scene-level audio without object-based representation; the primary comparison is among spatial audio methods, and spatial consistency is evaluated in Sec.~\ref{sec:poa_validation}.

\noindent\textbf{Audio Quality and Semantic Alignment.}
Among spatial audio methods, Scene2Sound achieves the best CLAP and, apart from the re-implemented SonoWorld (analyzed below), the best FAD and IB, while decomposing audio into per-object sources rather than generating a single scene-level waveform.
CLAP and IB score the mixture rendered at each viewpoint against the scene-level query.
The binaural statistics ($1{-}$IACC, ILD) confirm that every spatial method produces interaural cues; beyond that, their magnitudes reflect rendering strategy rather than quality. Methods that generate binaural waveforms directly can synthesize arbitrarily strong interaural differences, while our renderer mixes a non-directional ambient bed into the panned sources, which lowers the mixed-signal statistics. The per-source signals retain strong interaural cues before ambient mixing (supplementary material), and unlike a generated waveform can be re-spatialized at any listener pose.

\noindent\textbf{SonoWorld.} The audio-quality and semantic scores of the re-implemented SonoWorld partly reflect protocol effects. Its best pooled FAD reflects sparser mixes ($3.9$ placed sources vs.\ our $21.0$ instances per scene), and its weak binaural statistics trace to its MMAudio generator, a drop our own backend swap reproduces (Sec.~\ref{sec:sensitivity}).

\noindent\textbf{Real-World Scenes.}
On all $81$ real-world $360^\circ$ D-SAV360~\cite{BernalBerdun2023DSAV360} scenes, Scene2Sound attains the best FAD\textsuperscript{D}/FAD\textsuperscript{C} ($65.3$/$59.3$) among all methods, showing per-source audio quality transfers to 3DGS scenes generated from real captures. The re-implemented SonoWorld, applied to the same single capture position, is the closest spatial competitor on FAD but retains its weak semantic alignment (CLAP $0.080$). Table~\ref{tab:dsav360_results} reports the comparison against spatial audio methods (full comparison, incl.\ non-spatial baselines, in the supplementary material); the spatial signal here is analyzed in Sec.~\ref{sec:poa_validation}.
\input{sections/tables_dsav360.tex}

\subsection{Evaluating the Spatial Consistency of Audio}\label{sec:poa_validation}

\input{sections/tables_poa_validation.tex}

\noindent\textbf{Motivation.} No measure in Table~\ref{tab:main_results} captures spatial consistency under listener navigation: FAD, CLAP, and IB score single-viewpoint audio, and the binaural statistics describe cues within one rendering. Consistency has two faces: audio must \emph{respond} to listener motion consistent with scene geometry, and claimed sources must be \emph{supported} by visual evidence beyond the views used to place them. A smoothly varying but mislocalized renderer satisfies only the first, a well-placed but frozen soundscape only the second. We evaluate both axes.

\noindent\textbf{Listener axis.} \textit{Listener-Motion Consistency} (LMC) measures whether audio tracks listener displacement. For each triplet $(\mathbf{p}_a, \mathbf{p}_n, \mathbf{p}_f)$ with $d_{\mathrm{spatial}}(\mathbf{p}_a, \mathbf{p}_n) < d_{\mathrm{spatial}}(\mathbf{p}_a, \mathbf{p}_f)$, let $\Delta = d_{\mathrm{audio}}(a_{\mathbf{p}_a}, a_{\mathbf{p}_f}) - d_{\mathrm{audio}}(a_{\mathbf{p}_a}, a_{\mathbf{p}_n})$, where $d_{\mathrm{spatial}}$ is the Euclidean distance between listener positions, $a_{\mathbf{p}}$ is the audio at position $\mathbf{p}$, and $d_{\mathrm{audio}}$ the RMS distance between log-mel spectrograms. Over all valid triplets $\mathcal{T}$,
\begin{equation}
    \mathrm{LMC} = \frac{1}{|\mathcal{T}|} \sum_{(\mathbf{p}_a, \mathbf{p}_n, \mathbf{p}_f) \in \mathcal{T}} \psi(\Delta), \;
    \psi(\Delta) = \begin{cases} +1 & \Delta > \epsilon \\ 0 & |\Delta| \le \epsilon \\ -1 & \Delta < -\epsilon, \end{cases}
    \label{eq:lmc}
\end{equation}
with a small tie tolerance $\epsilon$, set relative to each evaluation set's distance scale (supplementary material); LMC ranges over $[-1,1]$ with chance level $0$. LMC alone cannot distinguish two very different ways of scoring near zero; we therefore report it alongside responsiveness $R$, the fraction of triplets whose audio changes at all ($|\Delta|>\epsilon$). A soundscape that transports one fixed waveform never changes, giving $(\mathrm{LMC},R)=(0,0)$; audio generated independently at each viewpoint changes constantly but without relation to listener geometry, giving $R\approx1$ with $\mathrm{LMC}\approx0$. A spatially consistent soundscape must do both: respond to motion ($R$ high) and respond in the direction the geometry dictates (LMC high).

\noindent\textbf{Grounding axis.} If the claimed sources are correctly grounded, each should reappear where the world itself shows the object, even in views that played no part in placing it. \textit{Cross-View Grounding Consistency} (CGC) turns this expectation into a held-out test. For each scene $s$ with views $\mathcal{V}_s$ and each fold view $v$, we re-run association and placement on the remaining views, project the resulting sources into $v$, and compare them with the object instances visible in $v$ (its segmentation masks). A projected source is supported when it falls inside a visible object, each object supports at most one source, and sources whose 3D position is hidden behind geometry in $v$ are excluded rather than penalized (the occlusion gate):
\begin{equation}
    \mathrm{CGC} = \frac{1}{|\mathcal{D}|}\sum_{s \in \mathcal{D}} \frac{1}{|\mathcal{V}_s|} \sum_{v \in \mathcal{V}_s} \frac{2\,H_{s,v}}{|P_{s,v}| + |M_{s,v}|},
    \label{eq:cgc}
\end{equation}
where $P_{s,v}$ are the sources placed without view $v$; $M_{s,v}$, the mask instances of $v$; $H_{s,v}$, the number of supported one-to-one pairs; and $\mathcal{D}$, the set of evaluated scenes. CGC is thus a held-out reprojection F1; its precision asks how many claimed sources are supported (penalizing duplicates and misplacements), and its recall asks how many visible objects are covered. The masks are a pseudo-reference, not ground truth; CGC ignores which audio a position carries; and since the VLM's event inventory is built once from all views, CGC measures held-out \emph{placement} consistency under a fixed inventory. Protocol and corruption tests are in the supplementary material.

\noindent\textbf{Method comparison.} Per-viewpoint baselines generate audio independently, giving responsive but inconsistent audio ($\mathrm{LMC}\approx0$, $R\approx1$) and no 3D sources (Table~\ref{tab:poa_validation}). SonoWorld places sources in 3D but scores below zero on the listener axis ($-0.214$, inverted audio-distance orderings) with weak held-out grounding (CGC $0.065$). Scene2Sound attains $\mathrm{LMC}=+0.283$, CGC $0.259$, and the highest held-out precision ($0.441$). The listener-axis contrasts are unambiguous (sign flips, zero-responsiveness nulls), and the CGC margin over SonoWorld is significant ($+0.19$; scene-level paired bootstrap, 95\% CI $[0.10, 0.29]$).

\noindent\textbf{Real-scene transfer.} On D-SAV360, the listener axis stays above the zero null under the single-capture-position regime, which precludes cross-position association; the transfer analysis is in the supplementary material.

\noindent\textbf{Validation.} On two real-world multi-position datasets with known coordinates (Real Acoustic Fields~\cite{chen2024raf}, dense RIRs; Replay-NVAS~\cite{chen2023novelview}, multi-microphone recordings), LMC is above the zero null (Table~\ref{tab:lmc_validation}), and Scene2Sound's $+0.283$ falls within this real-data band. Training ViGAS~\cite{chen2023novelview} on SoundSpaces-NVAS ($71$ clips), as a learning check, moves $(\mathrm{LMC}, R)$ from $(0,0)$ (untrained, viewpoint-invariant) to $(+0.455, 1.00)$ (clip-level paired Wilcoxon, $p{<}10^{-8}$): LMC rises with acquired spatial acoustic structure. Construct-validation controls (position corruptions for CGC, fixed-waveform/shuffle for LMC) behave as required; protocols are in the supplementary material.

\subsection{Qualitative Results}

Figure~\ref{fig:qualitative} illustrates instance association for a fan across four viewpoints: per-view masks are lifted to Gaussian sets and merged into a single instance whose re-projected 3D position falls within each original mask.

Figure~\ref{fig:instance_gallery} shows instance association across four diverse scenes spanning the types SoundscapePLY covers.
The bottom-right scene illustrates a failure case, where the segmentation model fails to ground any sound-emitting object and produces no point-source instances; the VLM still identifies ambient sound events, and Scene2Sound thus falls back to an ambient bed providing a plausible background atmosphere.
Video results of generated soundscapes, with navigable binaural audio, are available on the project page.

\subsection{Ablation Study}

\input{sections/tables_ablation.tex}

We ablate key components of Scene2Sound: Scene-level replaces per-source generation with one scene-wide clip from a VLM-produced prompt (cf.\ the SAO baseline in Table~\ref{tab:main_results}); Flat mix mixes all source audio into flat stereo without spatial rendering; Single viewpoint uses one viewpoint ($K{=}1$) instead of multiple; Random cameras samples $K{=}5$ cameras uniformly within the scene bounding box instead of automatic viewpoint selection; w/o instance association removes cross-view Gaussian set matching, treating each observation as a separate instance; and w/o ambient fold disables ambient bed separation, treating all sounds as point sources.
We report both spatial axes (LMC and CGC with its precision; Sec.~\ref{sec:poa_validation}) for each variant, along with the mean number of detected source labels ($N_{\mathrm{src}}$) and spatial instances ($N_{\mathrm{inst}}$) per scene; neither is a ground-truth quantity, since $N_{\mathrm{src}}$ counts VLM-identified sources and $N_{\mathrm{inst}}$ counts instances remaining after Gaussian set matching.

Table~\ref{tab:ablation} shows the results: Scene-level and Flat mix both reduce to fixed-waveform playback ($\mathrm{LMC}{=}0$, $R{=}0$), confirming that object-level decomposition and 3D spatial rendering are both essential for navigation-consistent soundscapes.
A single viewpoint detects fewer source labels ($N_{\mathrm{src}}{=}3.0$ vs.\ $3.4$) and fewer instances ($N_{\mathrm{inst}}{=}13.3$ vs.\ $21.0$), since sound events visible only from certain perspectives are missed; without informed camera placement (Random cameras), held-out grounding collapses to CGC $0.060$ on its 5-scene subset, though this ablates the entire camera-generation strategy; a milder raycast-random contrast is in the supplementary material.
Removing instance association leaves CGC F1 essentially unchanged ($0.2593$ vs.\ $0.2589$ before rounding) but triples the instances ($66.9$ vs.\ $21.0$) and shifts the operating point from precision toward recall ($0.302$/$0.316$ vs.\ our $0.441$/$0.229$); the precision difference is significant (paired 95\% CI $[0.05, 0.22]$), isolating the contribution of Gaussian set matching at a fixed inventory. Descriptively, placed sources rest on $2.72$ supporting views on average, and $65.6\%$ are observed from multiple views; most placed sources are thus corroborated by multiple views.
Without ambient fold, $N_{\mathrm{src}}$ rises to $5.2$ and the listener axis weakens ($+0.253 \to +0.217$ at the matched pre-normalization rendering; grounding nearly unchanged); the ambient/local distinction therefore matters for rendering rather than placement.

\subsection{Robustness Analyses}\label{sec:sensitivity}

\input{sections/tables_sensitivity.tex}

\noindent\textbf{Stage reliability.} A modular pipeline risks error accumulation; we therefore measure each stage's reliability on a manual annotation of all $124$ VLM proposals over the $24$ scenes: VLM source identification reaches $93.5\%$ proposal validity and $97.0\%$ element coverage; hallucinated/ungroundable sources are dropped when SAM3 grounding fails, and low-salience misses leave the ambient bed as the remaining background. Because sources are generated and rendered independently, an upstream error affects at most one source; full stage-wise diagnostics are in the supplementary material.

\noindent\textbf{Foundation-model sensitivity.} We vary the VLM and text-to-audio backend on a fixed 6-scene subset at the adopted configuration (Table~\ref{tab:sensitivity}; absolute values are not comparable to the 24-scene tables). Swapping Qwen2.5-VL-32B for the smaller 7B model cuts detected sources to a third ($3.2{\to}1.2$; one scene loses all point sources) and weakens both responsiveness ($R$ $1.00{\to}0.83$) and the listener axis ($+0.256{\to}+0.189$), while the audio-quality metrics barely register the impoverished soundscape. Replacing Stable Audio Open with TangoFlux or MMAudio keeps the listener axis positive throughout (${+}0.100$ to ${+}0.256$) and shifts mainly fidelity and semantics, since placement and geometry are held fixed: Scene2Sound is sensitive to its semantic front-end but robust to its interchangeable audio back-end. A 24-scene MMAudio re-run at the adopted prompts is consistent with this picture (FAD\textsuperscript{D} $97.9$, IB $0.091$); its mono output degrades the binaural statistics, and Stable Audio Open remains the normative backend.

\subsection{Subjective Evaluation}

\begin{figure}[t]
  \centering
  \includegraphics[width=0.95\linewidth]{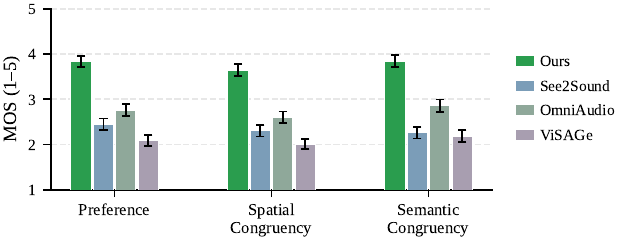}
  \caption{Subjective evaluation (MOS, 1--5 scale); error bars: standard error. The proposed method outperforms all baselines on all criteria.}
  \label{fig:subjective_eval}
\end{figure}

Quantitative evaluations use fixed-viewpoint settings for fair comparison with baselines lacking viewpoint-dependent rendering; to assess perceptual benefits under viewpoint changes, we ran a subjective evaluation with moving viewpoints (protocol and configuration note in the supplementary material): 19 self-reported normal-hearing participants rated the proposed method against three spatial audio baselines (See2Sound, OmniAudio, ViSAGe) on pre-rendered video clips with identical visual content, each evaluating four of 24 scenes over headphones on a 1--5 MOS scale for \textit{Preference} (overall quality), \textit{Spatial Congruency} (sound location vs.\ visual cues), and \textit{Semantic Congruency} (sound content vs.\ visual events).
All four conditions produce spatialized binaural audio (cf.\ the binaural statistics in Table~\ref{tab:main_results}); the method is therefore not trivially identifiable by spatial cues alone. A residual confound remains, since the conditions differ in how they track the moving viewpoint (ours adapts in full 6DoF via persistent 3D sources, OmniAudio follows head yaw only, ViSAGe re-infers audio from the recorded video), and a fully matched-cue study equalizing this is left as future work.

Figure~\ref{fig:subjective_eval} summarizes the results: Scene2Sound consistently outperforms all baselines on all three criteria (Preference $3.77$, Spatial Congruency $3.57$, Semantic Congruency $3.75$ vs.\ best-baseline MOS of $2.68$, $2.48$, $2.75$; means over all ratings, per-scene statistics in the supplementary material).
All nine pairwise comparisons (three baselines $\times$ three criteria) are significant under paired Wilcoxon signed-rank tests on participant--scene pairs with Bonferroni correction ($p<0.001$; effect sizes $r{=}0.55$--$0.76$).
The largest margins appear in Spatial Congruency, consistent with Scene2Sound's object-based binaural audio, which adapts dynamically as the listener moves, whereas the baselines lack explicit listener-position-dependent cues.

%% file: sections/tables_main_results.tex
\begin{table*}[tp]
\caption{Quantitative comparison on SoundscapePLY (24 scenes). \textbf{Bold}: best among spatial audio methods, the primary comparison targets; shading (\colorbox{red!15}{1st}, \colorbox{orange!20}{2nd}, \colorbox{yellow!25}{3rd}): rank across all methods (directional metrics only; the binaural columns are descriptive statistics, Sec.~\ref{sec:setup}).}
\label{tab:main_results}
\centering
\setlength{\tabcolsep}{3pt}
\begin{tblr}{
  width = \linewidth,
  colspec = {X[0.2,c] X[1.85,l] X[0.75,c] X[0.75,c] X[0.65,c] X[0.65,c] X[0.6,c] X[0.75,c]},
  rows = {font=\scriptsize},
  row{1-2} = {font=\scriptsize\bfseries},
  colsep = 2pt,
  rowsep = 1pt,
}
\hline
 & & \SetCell[c=2]{c} Audio Quality & & \SetCell[c=2]{c} Semantic & & \SetCell[c=2]{c} Binaural & \\
\cline{3-4} \cline{5-6} \cline{7-8}
 & Method & FAD\textsuperscript{D}$\downarrow$ & FAD\textsuperscript{C}$\downarrow$ & CLAP$\uparrow$ & IB$\uparrow$ & $1{-}$IACC & ILD \\
\hline
\SetCell[r=7]{c,m} \rotatebox{90}{\itshape Non-spatial} & AudioLDM2~\cite{liu2024audioldm2} & 111.5 & 117.0 & 0.195 & 0.123 & 0.000 & 0.00 \\
 & Tango2~\cite{majumder2024tango2} & 94.2 & 95.2 & 0.173 & \SetCell{bg=red!15} 0.146 & 0.000 & 0.00 \\
 & SAO~\cite{Evans2025StableAudio} & 94.2 & 81.2 & \SetCell{bg=yellow!25} 0.319 & 0.119 & 0.000 & 0.00 \\
 & MMAudio\textsuperscript{T}~\cite{cheng2025mmaudio} & \SetCell{bg=yellow!25} 87.8 & \SetCell{bg=orange!20} 77.2 & \SetCell{bg=red!15} 0.353 & 0.122 & 0.000 & 0.00 \\
 & MMAudio\textsuperscript{V}~\cite{cheng2025mmaudio} & 101.6 & 94.2 & 0.306 & 0.120 & 0.000 & 0.00 \\
 & Seeing\&Hearing~\cite{xing2024seeinghearing} & 100.7 & 102.2 & 0.137 & 0.116 & 0.000 & 0.00 \\
 & Im2Wav~\cite{sheffer2023im2wav} & 93.8 & 98.5 & 0.034 & 0.062 & 0.000 & 0.00 \\
\hline
\SetCell[r=4]{c,m} \rotatebox{90}{\itshape Spatial} & ViSAGe~\cite{kim2025visage} & 96.8 & 102.9 & 0.011 & 0.057 & 0.155 & 3.40 \\
 & See2Sound~\cite{Dagli2025SEE2SOUND} & 99.7 & 95.9 & 0.066 & 0.037 & 0.615 & 7.04 \\
 & OmniAudio~\cite{liu2025omniaudiogeneratingspatialaudio} & 94.7 & 90.5 & 0.205 & 0.110 & 0.458 & 6.74 \\
 & SonoWorld~\cite{sonoworld2026} (re-impl.) & \SetCell{bg=red!15} \textbf{79.3} & \SetCell{bg=red!15} \textbf{74.2} & 0.130 & \SetCell{bg=orange!20} \textbf{0.144} & 0.091 & 3.16 \\
\hline
 & Scene2Sound (Ours) & \SetCell{bg=orange!20} 83.8 & \SetCell{bg=yellow!25} 77.5 & \SetCell{bg=orange!20} \textbf{0.320} & \SetCell{bg=yellow!25} 0.128 & 0.254 & 6.24 \\
\hline
\end{tblr}
\end{table*}

%% file: sections/tables_dsav360.tex
\begin{table}[t]
\centering
\caption{
    \textbf{Quantitative comparison on D-SAV360} (81 real-world $360^\circ$ scenes), among spatial audio methods; full comparison incl. non-spatial baselines in the supplementary material.
    \textbf{Bold}: best (directional metrics; the binaural columns are descriptive statistics); FAD\textsuperscript{DS} uses D-SAV360's own recordings as the reference set.
}
\label{tab:dsav360_results}
\begin{tblr}{
  width = \linewidth,
  colspec = {X[2.1,l] X[0.7,c] X[0.7,c] X[0.7,c] X[0.9,c] X[0.6,c] X[0.75,c] X[0.6,c]},
  rows = {font=\scriptsize},
  row{1-2} = {font=\scriptsize\bfseries},
  colsep = 1.5pt,
  rowsep = 1pt,
}
\hline
 & \SetCell[c=3]{c} {Audio Quality} & & & \SetCell[c=2]{c} {Semantic} & & \SetCell[c=2]{c} {Binaural} & \\
\cline{2-4} \cline{5-6} \cline{7-8}
Method & {FAD\textsuperscript{D}$\downarrow$} & {FAD\textsuperscript{C}$\downarrow$} & {FAD\textsuperscript{DS}$\downarrow$} & {CLAP\,$\uparrow$} & {IB\,$\uparrow$} & {$1{-}$IACC} & {ILD} \\
\hline
ViSAGe~\cite{kim2025visage} & 73.3 & 77.0 & 75.0 & $-$0.004 & \textbf{0.141} & 0.291 & 4.48 \\
See2Sound~\cite{Dagli2025SEE2SOUND} & 83.2 & 76.2 & 100.1 & 0.058 & 0.053 & 0.616 & 7.41 \\
OmniAudio~\cite{liu2025omniaudiogeneratingspatialaudio} & 75.5 & 71.6 & \textbf{63.7} & 0.136 & 0.135 & 0.364 & 7.22 \\
SonoWorld~\cite{sonoworld2026} (re-impl.) & 67.6 & 64.2 & 73.6 & 0.080 & 0.129 & 0.577 & 8.65 \\
\hline
Scene2Sound (Ours) & \textbf{65.3} & \textbf{59.3} & 84.4 & \textbf{0.265} & 0.120 & 0.253 & 6.47 \\
\hline
\end{tblr}
\end{table}

%% file: sections/tables_poa_validation.tex
\begin{table}[t]
\centering
\caption{\textbf{Spatial consistency on both axes} (SoundscapePLY, 24 scenes). Listener axis: LMC $\in[-1,1]$ (chance $0$) with responsiveness $R$; grounding axis: CGC (Eq.~(\ref{eq:cgc})) with precision/recall. Best per column in bold; \textsuperscript{e}\,expected under independent per-viewpoint generation (not measured).}
\label{tab:poa_validation}
\footnotesize
\setlength{\tabcolsep}{2.5pt}
\begin{tabular}{lccccc}
\toprule
Method & LMC$\uparrow$ & $R$ & CGC$\uparrow$ & Prec.$\uparrow$ & Rec.$\uparrow$ \\
\midrule
per-viewpoint baselines (Table~\ref{tab:main_results}) & ${\approx}0$\textsuperscript{e} & ${\approx}1$\textsuperscript{e} & \textemdash{} & \textemdash{} & \textemdash{} \\
SonoWorld~\cite{sonoworld2026} (re-impl.) & $-0.214$ & $0.96$ & $0.065$ & $0.124$ & $0.063$ \\
Scene2Sound (Ours) & $\mathbf{+0.283}$ & $0.96$ & $\mathbf{0.259}$ & $\mathbf{0.441}$ & $\mathbf{0.229}$ \\
\bottomrule
\end{tabular}
\end{table}

\begin{table}[t]
\centering
\caption{\textbf{LMC validation.} Real multi-position recordings score well above the zero null, and training an acoustic-synthesis model moves LMC from exactly zero to strongly positive, supporting LMC as a measure of distance-dependent acoustic structure.}
\label{tab:lmc_validation}
\footnotesize
\setlength{\tabcolsep}{4pt}
\begin{tabular}{lcc}
\toprule
 & LMC$\uparrow$ & $R$ \\
\midrule
Real Acoustic Fields~\cite{chen2024raf} (dense RIRs) & $+0.310$ & $1.00$ \\
Replay-NVAS~\cite{chen2023novelview} (recordings) & $+0.207$ & $0.95$ \\
ViGAS~\cite{chen2023novelview}, random init.\ $\rightarrow$ trained & $0 \rightarrow +0.455$ & $0 \rightarrow 1.00$ \\
\bottomrule
\end{tabular}
\end{table}

%% file: sections/tables_ablation.tex
\begin{table*}[tp]
\caption{Ablation study on SoundscapePLY. Best in \textbf{bold}. LMC/CGC/Prec.\ defined in Sec.~\ref{sec:poa_validation}. ``\textemdash''\,=\,not applicable by construction (Scene-level/Flat mix: no 3D instances, fixed waveform, $\mathrm{LMC}{=}0$, $R{=}0$; Single-viewpoint: LMC needs $K\,{\geq}\,3$ positions). ``n.e.''\,=\,not evaluated (Random cameras targets grounding only; 5-scene subset). \textsuperscript{$\dagger$}\,original per-variant audio metrics; \textsuperscript{$\ddagger$}\,LMC re-scored at the adopted placement configuration with the variant's cached pre-adoption audio; only the Scene2Sound row uses the adopted audio prompts and gain normalization.}
\label{tab:ablation}
\centering
\setlength{\tabcolsep}{3pt}
\begin{tblr}{
  width = \linewidth,
  colspec = {X[2.0,l] X[0.75,c] X[0.65,c] X[0.65,c] X[0.75,c] Q[c,1pt] X[0.7,c] X[0.7,c] X[0.6,c] X[0.6,c]},
  rows = {font=\footnotesize},
  row{1} = {font=\footnotesize\bfseries},
  colsep = 2pt,
  rowsep = 1pt,
  vline{6} = {solid},
}
\hline
Variant & FAD$\downarrow$ & IB$\uparrow$ & ILD & LMC$\uparrow$ & & CGC$\uparrow$ & Prec.$\uparrow$ & $N_{\mathrm{src}}$ & $N_{\mathrm{inst}}$ \\
\hline
Scene2Sound & 83.8 & \textbf{0.128} & 6.24 & $\mathbf{+0.283}$ & & \textbf{0.259} & \textbf{0.441} & 3.4 & 21.0 \\
\hline
Scene-level & 90.6\textsuperscript{$\dagger$} & 0.110\textsuperscript{$\dagger$} & 5.49\textsuperscript{$\dagger$} & $0$ & & \textemdash & \textemdash & \textemdash & \textemdash \\
Flat mix & \textbf{81.4}\textsuperscript{$\dagger$} & 0.098\textsuperscript{$\dagger$} & 2.90\textsuperscript{$\dagger$} & $0$ & & \textemdash & \textemdash & \textemdash & \textemdash \\
\hline
Single viewpoint & 84.5\textsuperscript{$\dagger$} & 0.112\textsuperscript{$\dagger$} & 5.81\textsuperscript{$\dagger$} & \textemdash & & 0.187 & 0.322 & 3.0 & 13.3 \\
Random cameras & n.e. & n.e. & n.e. & n.e. & & 0.060 & 0.100 & n.e. & n.e. \\
w/o inst.\ assoc. & 84.3\textsuperscript{$\dagger$} & 0.114\textsuperscript{$\dagger$} & 7.31\textsuperscript{$\dagger$} & $+0.228$\textsuperscript{$\ddagger$} & & \textbf{0.259} & 0.302 & 3.4 & 66.9 \\
w/o Ambient fold & 82.8\textsuperscript{$\dagger$} & 0.122\textsuperscript{$\dagger$} & 6.43\textsuperscript{$\dagger$} & $+0.217$\textsuperscript{$\ddagger$} & & 0.256 & 0.424 & 5.2 & 21.4 \\
\hline
\end{tblr}
\end{table*}

%% file: sections/tables_sensitivity.tex
\begin{table*}[tp]
\centering
\caption{\textbf{Sensitivity to foundation-model choice} on a fixed 6-scene SoundscapePLY subset at the adopted configuration (not comparable to the 24-scene tables): (a) VLM swap, (b) text-to-audio backend swap. Best per column in bold in (b).}
\label{tab:sensitivity}
\begin{minipage}[t]{0.54\textwidth}
\centering
\footnotesize
(a) VLM swap (T2A = Stable Audio Open)\\[2pt]
\setlength{\tabcolsep}{4pt}
\begin{tabular}{lcccccc}
\toprule
VLM & \#Src & LMC$\uparrow$ & $R$ & FAD$\downarrow$ & CLAP$\uparrow$ & IB$\uparrow$ \\
\midrule
Qwen2.5-VL-32B-AWQ & 3.2 & $+0.256$ & 1.00 & 106.6 & 0.349 & 0.126 \\
Qwen2.5-VL-7B & 1.2 & $+0.189$ & 0.83 & 103.4 & 0.308 & 0.115 \\
\bottomrule
\end{tabular}
\end{minipage}\hfill
\begin{minipage}[t]{0.42\textwidth}
\centering
\footnotesize
(b) T2A swap (VLM = Qwen2.5-VL-32B-AWQ)\\[2pt]
\setlength{\tabcolsep}{4pt}
\begin{tabular}{lcccc}
\toprule
T2A backend & FAD$\downarrow$ & CLAP$\uparrow$ & IB$\uparrow$ & LMC$\uparrow$ \\
\midrule
Stable Audio Open & 106.6 & \textbf{0.349} & \textbf{0.126} & $\mathbf{+0.256}$ \\
TangoFlux & 109.1 & 0.270 & 0.066 & $+0.100$ \\
MMAudio & \textbf{102.8} & 0.334 & 0.124 & $+0.133$ \\
\bottomrule
\end{tabular}
\end{minipage}
\end{table*}

%% file: sections/discussion.tex
\section{Discussion}\label{sec:discussion}

Despite these results, several challenges remain. Because our setting is a static 3DGS, which supplies appearance, geometry, and a persistent primitive index but no dynamic state, sounds that depend on motion or interaction (a fan spinning up, collisions) reflect the object's typical sound rather than its instantaneous state; coupling with dynamic or physics-aware scene representations is a natural next step. As a training-free pipeline orchestrating off-the-shelf foundation models, Scene2Sound's final quality inherits the limitations of each component, and errors cascade through the pipeline: a missed VLM detection, inaccurate segmentation, or poor audio synthesis each propagates downstream. The cascade is most critical at viewpoint selection, where inadequate camera placement (e.g., when the hollow-structure assumption fails on large or complex scenes) makes downstream modules miss sources or fail cross-view association. Our method also does not model acoustic propagation effects such as reverberation or occlusion.
Evaluation data are a further limitation: our real-scene check is confined to the single-view regime because, to our knowledge, no public dataset pairs multi-view real captures with environmental soundscapes and multi-instance source annotations, and assembling such a benchmark is an open need for this task. Relatedly, LMC and CGC test motion response and geometric support separately; neither verifies that the right sound is bound to the right instance, which would require annotated audio--object correspondences.
Promising future directions include geometry-based RIR estimation or learned acoustic fields~\cite{liang2023AVNeRF,bhosale2024avgs,brunetto2025neraf} to add propagation effects on top of our object-based representation, and joint optimization or iterative feedback across modules to mitigate error accumulation.

\section{Conclusion}\label{sec:conclusion}

We studied soundscape generation for a given 3DGS world through auditory grounding and proposed Scene2Sound, a training-free framework that anchors generated sounds to view-consistent 3D instances via Gaussian set matching and renders them with a standard object-based audio engine. Across SoundscapePLY and 3DGS scenes generated from real-world $360^\circ$ captures, Scene2Sound preserves per-viewpoint audio quality while remaining spatially consistent on both evaluation axes, and a user study confirms the perceptual benefit. The proposed metrics and the SoundscapePLY testbed support follow-up research on extending visual 3DGS worlds into multimodal world simulations.

%% file: sections/supp/A1_DatasetDetails.tex
\label{sec:supp_dataset}

SoundscapePLY contains 24 3DGS scenes generated with Marble~\cite{marble}, each comprising a structured text prompt, a concept image, and the resulting 3D Gaussian splats, spanning urban, natural, indoor, historical, and stylized visual domains with distinct soundscape characteristics.

\paragraph{Generation pipeline.}
Each prompt specifies scene identity (world type, canonical viewpoint), style, mood, layout (interior/exterior, spatial organization, depth cues), soundscape (ambient background and localized object sounds), and materials/lighting. We generate a concept image per prompt with FLUX.2 [klein] 9B~\cite{blackforestlabs2026flux2klein9b} and feed it to Marble to produce the final 3DGS scene.

\paragraph{Data availability.}
All prompts, concept images, and 3DGS scenes were produced by the authors under Marble's paid subscription plan, which permits redistribution; the full dataset will be released upon publication.

%% file: sections/supp/A2_ViewpointSelection.tex
\label{sec:supp_viewpoint}

We select $K$ panoramic viewpoints per scene by greedily maximizing the main paper's objective $\mathrm{Coverage}(\mathcal{K})$, denoted $\mathrm{Coverage}_q$ below: candidates are cast via a Fibonacci-lattice raycast from a robust scene center toward the density shell and scored by visible-Gaussian coverage weighted by CLIP-IQA~\cite{wang2023clipiqa} and inverse viewing distance. Greedy selection reaches $0.772$ raw coverage and $0.838$ CLIP-IQA on average.

\paragraph{Evaluation.}
Table~\ref{tab:supp_viewpoint_comparison} compares candidate-sampling strategies and selection algorithms at $K{=}5$ over the 24 SoundscapePLY scenes. Raycast sampling dominates ($\mathrm{Coverage}_q$ $0.52$--$0.59$ vs.\ $\le 0.141$ for non-raycast) and greedy selection improves visual quality over random. Running the full pipeline with random camera placement drops held-out grounding to CGC $0.060$ on its 5-scene subset (main-paper ablation).

\begin{table}[t]
\centering
\caption{
    \textbf{Viewpoint selection comparison} on SoundscapePLY (24 scenes, $K{=}5$).
    $\mathrm{Coverage}_q$ is the primary metric combining Gaussian coverage with visual quality. Raw Cov.: unweighted visible-Gaussian coverage.
}
\label{tab:supp_viewpoint_comparison}
\scriptsize
\setlength{\tabcolsep}{3pt}
\begin{tblr}{
  colspec = {llccc},
  colsep = 3pt,
  row{1} = {font=\bfseries},
  hline{1} = {0.08em},
  hline{2} = {0.05em},
  hline{4} = {2-5}{dashed},
  hline{6} = {0.05em},
  hline{9} = {0.08em},
}
  Sampling & Selection & $\mathrm{Coverage}_q\!\uparrow$ & Raw Cov.$\uparrow$ & CLIP-IQA$\uparrow$ \\
  AABB Uniform & Random & 0.084 \tiny{$\pm$0.062} & 0.334 \tiny{$\pm$0.151} & 0.469 \tiny{$\pm$0.098} \\
  AABB Uniform & Greedy & 0.141 \tiny{$\pm$0.061} & 0.542 \tiny{$\pm$0.202} & 0.698 \tiny{$\pm$0.115} \\
  Raycast & Random & 0.518 \tiny{$\pm$0.153} & 0.738 \tiny{$\pm$0.080} & 0.498 \tiny{$\pm$0.082} \\
  \textbf{Raycast (Ours)} & \textbf{Greedy} & \textbf{0.587} \tiny{$\pm$0.141} & \textbf{0.772} \tiny{$\pm$0.078} & 0.838 \tiny{$\pm$0.068} \\
  FPS Floor & Greedy & 0.087 \tiny{$\pm$0.091} & 0.284 \tiny{$\pm$0.294} & 0.762 \tiny{$\pm$0.075} \\
  Density Peak & Greedy & 0.124 \tiny{$\pm$0.087} & 0.549 \tiny{$\pm$0.159} & 0.656 \tiny{$\pm$0.098} \\
  Raycast (Fixed) & Greedy & 0.583 \tiny{$\pm$0.141} & 0.771 \tiny{$\pm$0.080} & \textbf{0.842} \tiny{$\pm$0.066} \\
\end{tblr}
\end{table}

%% file: sections/supp/A4_AudioPlacement.tex
\label{sec:supp_placement}

\paragraph{Depth-Consistency Gate for Background Removal.}
Tile-based rendering metadata returns all Gaussians contributing to tiles that overlap the segmentation mask, including background Gaussians (\eg, walls behind the target). We remove them with a one-sided, scale-relative depth-consistency gate: Gaussians farther from the camera than $\alpha_d \cdot d_{\mathrm{med}}$, with $d_{\mathrm{med}}$ the median rendered depth inside the mask, are discarded. The gate is one-sided (tile overflow only adds primitives \emph{behind} the object) and scale-relative (the tolerated spread grows with distance); $\alpha_d{=}1.5$ is a fixed tolerance, and an emptied set falls back to the unfiltered one. The resulting set $\mathcal{G}_{k,m}$ and its weighted centroid then reflect the target object rather than surrounding structure.

\paragraph{Sensitivity to the Jaccard Threshold.}
Table~\ref{tab:supp_jmin_sweep} sweeps the matching threshold $J_{\min}$ on all 24 scenes under the held-out CGC protocol of the main paper, with other parameters at their adopted values. The adopted $J_{\min}{=}0.15$ balances held-out F1, precision, and the listener axis; an earlier version of this work selected $0.05$ under its in-sample reprojection score (legacy ILA; Sec.~\ref{sec:supp_spatial_metrics}). The final configuration ($J_{\min}{=}0.15$, no instance cap, scene-context audio prompts) comes from a pre-specified search with paired promotion gates on a fixed 8-scene panel; a final rendering revision (per-label instance-energy normalization, ambient bed $0.4$), adopted after an author listening review, improves every audio metric and the listener axis. The sweep's LMC row predates this revision (adopted column $+0.253$ vs.\ $+0.283$ in Table~III). All reported results use the final configuration except the subjective study, which states its earlier one.

\begin{table}[t]
\centering
\caption{Sensitivity of instance association to the Jaccard threshold $J_{\min}$ (24 scenes, held-out CGC protocol). The adopted value is marked in \textbf{bold}.}
\label{tab:supp_jmin_sweep}
\scriptsize
\setlength{\tabcolsep}{3pt}
\begin{tabular}{lcccccc}
\toprule
$J_{\min}$ & 0.01 & 0.05 & 0.10 & \textbf{0.15} & 0.20 & 0.50 \\
\midrule
CGC $\uparrow$ & 0.159 & 0.224 & 0.251 & \textbf{0.259} & 0.265 & 0.286 \\
Prec.\ $\uparrow$ & 0.320 & 0.410 & 0.440 & \textbf{0.441} & 0.443 & 0.396 \\
LMC $\uparrow$ & $+$0.206 & $+$0.211 & $+$0.236 & $\mathbf{+0.253}$ & $+$0.247 & $+$0.244 \\
\bottomrule
\end{tabular}
\end{table}

\paragraph{Co-Located Same-Class Instances.}
A potential failure mode of Jaccard-based matching is over-merging distinct same-class instances standing close together. Across the five scenes with same-label multi-instance sources ($12$ sources, $2$--$3$ instances each), every cross-camera observation pair belonging to \emph{different} instances of the same label was checked: $0$ of $16{,}284$ pairs exceeded the adopted $J_{\min}$ (no over-merge), including instances separated by as little as $0.23$\,m. This is structural: the segmentation model already separates same-class instances per image, thus Gaussian set matching resolves only cross-view correspondences of the same instance.

%% file: sections/supp/A5_VLMPromptDesign.tex
\label{sec:supp_vlm_prompt}

The system prompt sent to Qwen2.5-VL-32B for scene understanding instructs the model to design realistic spatial audio for the scene and to emit only JSON matching a fixed schema; the user message supplies $K{=}5$ panoramic images and requests up to 6 sound sources. Each source carries a \texttt{source\_type} (\texttt{point} vs.\ \texttt{ambient}), a short \texttt{sound\_label}, an \texttt{audio\_prompt} (temporal behavior and acoustic texture), \texttt{visible\_in\_views}, per-view \texttt{grounding\_queries} (1--3 word segmentable nouns for SAM3), and \texttt{estimated\_size}/\texttt{estimated\_loudness}. Point sources must be grounded to a specific visible object; ambient sources set \texttt{visible\_in\_views} to \texttt{[]} and are folded into the ambient bed. Core design rules require atmosphere-first source selection from the place, season, and weather; active sounds only, thus static materials (wood, stone, glass) stay silent; and grounding queries that name a specific, view-unique, segmentable object (\eg, ``ceiling fan'', not ``landscape''). The \texttt{estimated\_size} field sets each source's reference distance and rolloff factor for distance attenuation, and \texttt{estimated\_loudness} sets its base gain. Common violations of these rules (duplicated or overly generic queries, human-presence descriptions) are illustrated in Section~\ref{sec:supp_qualitative}; the complete prompt, few-shot exemplars, and JSON schema ship with the code release.

\paragraph{Reliability of VLM Source Identification.}
We manually verified all $124$ sources proposed by the VLM on the 24 SoundscapePLY scenes against ground-truth key elements. Proposal validity is $93.5\%$: $8$ of $124$ proposals are hallucinations, following an urban prior (\eg, ``traffic noise'' in a flooded ruin). Element coverage is $97.0\%$: $5$ of $168$ ground-truth elements lack a proposal, largely absorbed by the ambient bed. A proposal may cover several elements, thus the two rates use different matching units. Additionally, $25/124$ ($20.2\%$) proposed sources fail SAM3 grounding in all visible views: grounding correctly filters the hallucinations above and other non-segmentable proposals, but also misses small or fine-detail objects (e.g., area-noun queries such as ``shoreline''), leaving the ambient bed as background. The dropout is localized to the affected source; surviving sources are unaffected.

%% file: sections/supp/A6_POAMetricValidation.tex
\label{sec:supp_spatial_metrics}

\paragraph{LMC protocol.}
$d_{\mathrm{audio}}$ is the RMS distance between log-mel spectrograms ($128$ mel bins; $92.9$\,ms window with $23.2$\,ms hop at $22.05$\,kHz) of mono-downmixed renders. The tie tolerance is $\epsilon = 10^{-6}$ times the median pairwise audio distance in the evaluation set; responsiveness $R$ is the fraction of triplets with $|\Delta| > \epsilon$. LMC relates to the earlier POA formulation as $\mathrm{LMC} = 2\,\mathrm{POA} - 1$ on tie-free data, with ties scored $0$ deterministically.

\paragraph{CGC protocol.}
For each scene ($K{=}5$ rig views) and fold view $v$, we remove all observations from $v$ before instance association, re-run clustering and position estimation on the remaining views, and project the resulting instances into $v$. A projected instance is evaluable only if unoccluded (camera distance at most $1.1\times$ the rendered 3DGS depth at its pixel). Evaluable predictions are matched one-to-one to the held-out view's mask instances via maximum bipartite matching on point-in-mask membership (same frozen mask bank for all methods). Per-view F1 is $2H/(|P|+|M|)$, averaged over evaluable views then scenes, with scene-level bootstrap ($1000$ resamples) for confidence intervals, and paired method contrasts use $10{,}000$ scene-level resamples over the $23$ mutually evaluable scenes; single-panorama methods use the rig views never used in their construction. CGC is an unlabeled-set metric measuring held-out geometric support, coverage, and duplicate control, invariant to which audio or label a position carries. The VLM event inventory and grounding queries are built once from all views (not re-run per fold), thus CGC evaluates held-out placement consistency conditioned on that fixed inventory; views with neither predictions nor masks are non-evaluable, and occlusion-gated predictions leave the precision pool.

\paragraph{Construct validation.}
Both metrics were checked against controlled corruptions of our own outputs (run at the pre-adoption configuration; the gate validates metric behavior, not the method): rotating all source positions $90^\circ$ collapses CGC ($0.143 \to 0.022$); duplicating every source lowers precision ($0.350 \to 0.188$); deleting half the sources lowers recall ($0.121 \to 0.066$); permuting ownership of a fixed position set leaves CGC unchanged, as required by its unlabeled-set definition; copying one waveform to all listener positions yields $(\mathrm{LMC}, R) = (0, 0)$ exactly; shuffling listener--audio assignments yields $\mathrm{LMC} \approx 0$ with $R$ near $1$.

\paragraph{Real-recording and learning validation.}
RAF~\cite{chen2024raf} (dense RIRs, 100 groups of 8 receivers/transmitter) and Replay-NVAS~\cite{chen2023novelview} (natural 8-mic recordings, 41 scenes) both use known microphone coordinates and no renderer of ours, giving $\mathrm{LMC} = +0.310$ and $+0.207$, above the zero null (main paper, Table~IV). For the learning check, we render the 71-clip SoundSpaces-NVAS protocol with the released ViGAS checkpoint and with random initialization (same clips, positions, script; bit-deterministic inference). The untrained network outputs near-constant, viewpoint-invariant audio, thus every triplet ties, $(\mathrm{LMC}, R) = (0, 0)$; the trained model reaches $(+0.455, 1.00)$ (paired Wilcoxon $p{=}9.0{\times}10^{-9}$; scene-cluster bootstrap over 8 scenes, $95\%$ CI $[+0.367, +0.559]$).

\paragraph{Legacy results.}
Earlier versions reported the listener axis as POA and a mask-based score (ILA). POA maps to LMC via the relation above; ILA is retired since it verified predictions against the masks used to construct them and did not penalize misses, now corrected by CGC's held-out protocol.

%% file: sections/supp/A7_SubjectiveEvaluation.tex
\label{sec:supp_subjective}

\noindent\textbf{Participants and environment.}
19 participants (predominantly graduate students in their 20s, self-reported normal hearing) evaluated remotely on personal headphones, reflecting the realistic personal-device use case for interactive 3DGS exploration.

\noindent\textbf{Stimuli.}
We used pre-rendered 10-second video clips with author-designed camera trajectories (translation, rotation) to exercise 6DoF cues, with strict visual parity so rating differences are attributable to audio alone; stimuli were produced at an earlier pipeline configuration (placement cap $3$, $J_{\min}{=}0.05$, earlier audio-prompt template), while the object-based rendering mechanism evaluated is unchanged. Four conditions were generated per scene: (1)~\textbf{Scene2Sound (Ours)}, object-based binaural audio via HRTF spatialization adapting to listener position/orientation in 6DoF; (2)~\textbf{See2Sound~\cite{Dagli2025SEE2SOUND}}, 5.1-channel surround from a single image downmixed to stereo (ITU coefficients), a fixed layout that cannot adapt to viewpoint; (3)~\textbf{OmniAudio~\cite{liu2025omniaudiogeneratingspatialaudio}}, FOA from a fixed panoramic image with per-frame yaw rotation, supporting head rotation but not translation; and (4)~\textbf{ViSAGe~\cite{kim2025visage}}, FOA re-inferred from the recorded moving-viewpoint video with the same yaw-rotation and binaural decoding, testing whether video-conditioned generation captures viewpoint-dependent cues. All conditions shared identical visual content, loudness-normalized to $-23$\,LUFS (EBU~R128).

\noindent\textbf{Procedure.}
Each participant rated four assigned scenes on all four conditions using a 1--5 MOS scale for Preference, Spatial Congruency, and Semantic Congruency, order randomized per scene. Pairwise Wilcoxon tests ($n{=}76$/condition) show Scene2Sound outperforms every baseline on all three criteria ($p<0.001$, Bonferroni-corrected; $r=0.55$--$0.76$).

\noindent\textbf{Per-scene variability.}
The standard deviation of per-scene mean MOS over the 24 scenes is $0.95$--$1.10$ for Scene2Sound (Preference $3.73\pm1.02$, Spatial Congruency $3.50\pm0.95$, Semantic Congruency $3.73\pm1.10$; unweighted means over per-scene means, hence slightly below the main paper's rating-level means) and $0.69$--$0.96$ for the baselines, reflecting Scene2Sound's dependence on upstream grounding: well-grounded scenes approach ceiling ratings, while grounding failures pull individual scenes down. The pairwise tests above use matched participant--scene pairs, thus significance already accounts for this variation.

\noindent\textbf{Limitations of the protocol.}
Scene2Sound is the only condition rendering object-dependent distance attenuation, thus part of the subjective gain may reflect the availability of 6DoF distance cues, a target capability rather than an artifact, though this study does not isolate it from method identity. Since all four conditions are spatialized binaural (cf.\ the binaural statistics in Table~I of the main paper), the method is not identifiable merely by stereo cues; a matched study toggling distance attenuation within the same renderer would isolate this factor, left to future work.

\paragraph{Binaural cue analysis.}
The moderate ILD of our final output (Table~I of the main paper: $6.24$\,dB) is a design consequence of the ambient/point-source decomposition, not weak spatialization. Before ambient mixing, per-source object audio exhibits strong binaural cues: the median per-scene ILD over the 24 scenes is $7.32$\,dB (same STFT protocol as Table~I), comparable to See2Sound's mixed output ($7.04$\,dB). Mixing in the deliberately omnidirectional ambient bed then softens the final value.

%% file: sections/supp/A8_QualitativeResults.tex
\label{sec:supp_qualitative}

Across the 24 SoundscapePLY scenes, the VLM reliably grounds object-specific point sources in visually distinctive settings (\eg, fireplace, candle, and lamp sounds in a Victorian study; fountain and wind-chime sounds in a marketplace). Two failure patterns recur: overly generic grounding queries (\eg, ``underwater'' for a coral reef scene) name a non-segmentable area rather than an object, thus SAM correctly rejects the query; and scene descriptions occasionally evoke a nearby individual (\eg, ``soft footsteps of visitors''), violating the human-presence exclusion rule of Section~\ref{sec:supp_vlm_prompt}. The complete qualitative gallery is provided as multimedia material.

%% file: sections/supp/A11_DSAV360Protocol.tex
\paragraph{Dataset and Setup.}
We use D-SAV360~\cite{BernalBerdun2023DSAV360}, a curated collection of 85 real-world $360^\circ$ video scenes with diverse soundscapes (urban streets, parks, restaurants, train stations). From each scene we extract a representative equirectangular frame and reconstruct a 3DGS with DreamScene360~\cite{zhou2024dreamscene360}, obtaining 81 scenes across five environment categories after excluding four failed reconstructions. Each scene originates from a single fixed-position $360^\circ$ image, thus we skip viewpoint selection and observe from the original camera position alone, rendering three yaw-rotated panoramas ($0^\circ/120^\circ/240^\circ$) to reduce equirectangular distortion; Gaussian set matching associates detections only across these same-position renderings, thus no cross-position association is possible. All other stages are unchanged; we compare against nine of the eleven main-paper baselines (SAO and SonoWorld are evaluated on SoundscapePLY only).

\paragraph{Metrics.}
Audio quality uses dataset-level FAD (PANNs) against DCASE, Clotho, and D-SAV360's ambisonic recordings downmixed to mono; semantic alignment uses ImageBind and CLAP; binaural statistics use interaural decorrelation ($1{-}$IACC) and mean absolute ILD (metric references in the main paper). For the listener axis (LMC) we sample $15$ listener positions per scene via the same Fibonacci-sphere sampling as viewpoint candidate generation and evaluate over all valid triplets. LMC is reported for Scene2Sound only, since per-viewpoint baselines generate audio independently at each viewpoint; as a sanity check, ambient-only scenes yield $\mathrm{LMC}{\,=\,}0$ exactly.

\paragraph{Grounding Statistics and Listener-Axis Transfer.}
The VLM detected 4.8 source labels per scene on average, 4.2 grounded via SAM3. At least one label obtained a mask in 71 of 81 scenes, and 46 scenes yielded at least one anchored point source; the single-position setting limits grounding, since objects unseen from the capture position cannot be recovered from elsewhere. LMC is $+0.068$ over all $81$ scenes and $+0.120$ on the $46$ scenes with grounded point sources, above the zero null (one-sample $t$-test, $p\!<\!10^{-6}$) but below the generated-scene $+0.283$; the $35$ scenes without grounded sources score exactly $0$. Closing the gap to the generated-scene setting needs multi-position real captures, absent from public soundscape datasets. Table~\ref{tab:supp_dsav360_full} adds six non-spatial baselines omitted from the main paper's table.

\begin{table*}[tp]
\centering
\caption{
    \textbf{Full D-SAV360 comparison including non-spatial baselines} (81 real-world $360^\circ$ scenes).
    \textbf{Bold}: best among spatial audio methods; shading (\colorbox{red!15}{1st}, \colorbox{orange!20}{2nd}, \colorbox{yellow!25}{3rd}): rank across all methods (directional metrics only; the binaural columns are descriptive statistics).
    FAD reference sets: D\,=\,DCASE, C\,=\,Clotho, DS\,=\,D-SAV360.
}
\label{tab:supp_dsav360_full}
\begin{tblr}{
  width = \textwidth,
  colspec = {X[0.2,c] X[2.2,l] X[0.8,c] X[0.8,c] X[0.8,c] X[0.8,c] X[0.7,c] X[0.7,c] X[0.7,c]},
  rows = {font=\tiny},
  row{1-2} = {font=\tiny\bfseries},
  colsep = 2pt,
  rowsep = 0.5pt,
}
\hline
 & & \SetCell[c=3]{c} {Audio Quality (FAD\,$\downarrow$)} & & & \SetCell[c=2]{c} {Semantic} & & \SetCell[c=2]{c} {Binaural} & \\
\cline{3-5} \cline{6-7} \cline{8-9}
 & Method & {FAD\textsuperscript{D}} & {FAD\textsuperscript{C}} & {FAD\textsuperscript{DS}} & {CLAP\,$\uparrow$} & {IB\,$\uparrow$} & {$1{-}$IACC} & {ILD} \\
\hline
\SetCell[r=6]{c,m} \rotatebox{90}{\itshape Non-spatial} & AudioLDM2~\cite{liu2024audioldm2} & 73.5 & 67.4 & 84.6 & \SetCell{bg=yellow!25} 0.301 & 0.102 & 0.000 & 0.00 \\
 & Tango2~\cite{majumder2024tango2} & 72.2 & 73.4 & \SetCell{bg=orange!20} 67.4 & 0.251 & \SetCell{bg=yellow!25} 0.153 & 0.000 & 0.00 \\
 & MMAudio\textsuperscript{T}~\cite{cheng2025mmaudio} & 74.1 & \SetCell{bg=orange!20} 63.2 & 82.7 & \SetCell{bg=orange!20} 0.410 & \SetCell{bg=orange!20} 0.163 & 0.000 & 0.00 \\
 & MMAudio\textsuperscript{V}~\cite{cheng2025mmaudio} & 78.3 & 66.8 & 80.4 & \SetCell{bg=red!15} 0.424 & \SetCell{bg=red!15} 0.183 & 0.000 & 0.00 \\
 & Seeing\&Hearing~\cite{xing2024seeinghearing} & 82.2 & 75.6 & 93.6 & 0.252 & 0.117 & 0.000 & 0.00 \\
 & Im2Wav~\cite{sheffer2023im2wav} & \SetCell{bg=yellow!25} 69.2 & 71.2 & 80.1 & 0.074 & 0.148 & 0.000 & 0.00 \\
\hline
\SetCell[r=4]{c,m} \rotatebox{90}{\itshape Spatial} & ViSAGe~\cite{kim2025visage} & 73.3 & 77.0 & 75.0 & $-$0.004 & \textbf{0.141} & 0.291 & 4.48 \\
 & See2Sound~\cite{Dagli2025SEE2SOUND} & 83.2 & 76.2 & 100.1 & 0.058 & 0.053 & 0.616 & 7.41 \\
 & OmniAudio~\cite{liu2025omniaudiogeneratingspatialaudio} & 75.5 & 71.6 & \SetCell{bg=red!15} \textbf{63.7} & 0.136 & 0.135 & 0.364 & 7.22 \\
 & SonoWorld~\cite{sonoworld2026} (re-impl.) & \SetCell{bg=orange!20} 67.6 & \SetCell{bg=yellow!25} 64.2 & \SetCell{bg=yellow!25} 73.6 & 0.080 & 0.129 & 0.577 & 8.65 \\
\hline
 & Scene2Sound (Ours) & \SetCell{bg=red!15} \textbf{65.3} & \SetCell{bg=red!15} \textbf{59.3} & 84.4 & \textbf{0.265} & 0.120 & 0.253 & 6.47 \\
\hline
\end{tblr}
\end{table*}

%% file: sections/supp/A12_SonoWorldReimpl.tex
The SonoWorld baseline re-implements the pipeline \emph{shape} of SonoWorld from its released code, instantiated on our 3DGS worlds under module parity.
Kept: the single-panorama regime ($K{=}1$, the scene-center view our viewpoint selection visits first), its scene-understanding prompt (verbatim), its depth-weighted mask-lifting formula (\textsf{mask\_concentration}, verbatim), and its MMAudio generator.
Substituted where their component is closed or unreleased: GPT-4.1 $\rightarrow$ Qwen2.5-VL-32B (same VLM as ours, their prompt); MoGe depth $\rightarrow$ the scene's rendered depth; their unreleased renderer $\rightarrow$ our binaural engine at the same listener positions.
Over the 24 scenes ($\approx$418\,s/scene), the VLM proposed $136$ foreground sources, of which SAM3 grounded $93$ ($68.4\%$; $3.9$ placed/scene); held-out grounding follows the same CGC protocol as all other methods.
On D-SAV360 the same re-implementation runs unchanged on the dataset's single capture position, with depth from the same 3DGS reconstructions; in $4$ of $81$ scenes SAM3 grounds no source, leaving only the ambient bed (counted in the aggregate).

%% file: sections/tables_error_budget.tex
\begin{table}[t]
\centering
\caption{\textbf{Stage-wise diagnostics.} Measured reliability of each pipeline stage and the mechanism that limits its errors.}
\label{tab:error_budget}
\begin{tblr}{
  width = \linewidth,
  colspec = {X[1.1,l] X[1.6,l] X[2.6,l]},
  rows = {font=\tiny},
  row{1} = {font=\tiny\bfseries},
  colsep = 2pt,
  rowsep = 0.5pt,
}
\hline
Stage & Reliability & Failure mode $\rightarrow$ containment \\
\hline
VLM source detection & Proposal validity $93.5\%$, element coverage $97.0\%$ & $8/124$ hallucinations (urban prior) $\rightarrow$ rejected downstream by grounding; $5/168$ missed elements $\rightarrow$ absorbed by the ambient bed \\
SAM grounding & $99/124$ grounded ($20.2\%$ dropout) & ungroundable queries (small objects, area nouns, hallucinations) $\rightarrow$ dropped independently; surviving sources unaffected \\
GSM instance association & held-out grounding CGC $0.259$, precision $0.441$ (main paper, Tables~III and~V) & same-class over-merge $\rightarrow$ $0/16{,}284$ occurrences at the adopted threshold (Sec.~\ref{sec:supp_placement}) \\
Audio generation & competitive FAD among spatial methods (Table~I) & per-source artifacts $\rightarrow$ per-source independence; ambient bed as scene-level fallback \\
Rendering / listener axis & LMC $+0.283$ (generated), $+0.068$ (real) & deterministic re-rendering given instance positions \\
\hline
\end{tblr}
\end{table}

%% file: sections/supp/A9_RuntimeAnalysis.tex
\label{sec:supp_runtime}

Table~\ref{tab:supp_runtime_breakdown} reports the per-stage wall-clock breakdown over all 24 SoundscapePLY scenes on a single NVIDIA RTX A6000 (48\,GB).
VLM scene analysis dominates ($112$\,s, $50\%$; Qwen2.5-VL-32B, AWQ-quantized), followed by per-source audio generation ($59.7$\,s, $27\%$; $\approx$$12$\,s/source). The geometry-dependent stages (panoramic rendering, tile-based Gaussian voting, Jaccard clustering) take under $40$\,s combined, thus the 3DGS-side pipeline is lightweight.

\begin{table}[t]
\centering
\caption{Per-stage inference time (mean $\pm$ std over 24 scenes; NVIDIA RTX A6000).}
\label{tab:supp_runtime_breakdown}
\scriptsize
\setlength{\tabcolsep}{3pt}
\begin{tabular}{lrr}
\toprule
Stage & Time (s) & \% of total \\
\midrule
Rendering (3DGS $\to$ panoramas) & $21.3 \pm 0.3$ & 10\% \\
VLM scene analysis & $112.0 \pm 22.1$ & 50\% \\
SAM grounding & $12.7 \pm 0.8$ & 6\% \\
Gaussian voting (tile metadata) & $16.0 \pm 18.3$ & 7\% \\
Clustering (Jaccard / Union-Find) & $0.5 \pm 0.0$ & $<$1\% \\
Audio generation (SAO $\times N_{\mathrm{src}}$) & $59.7 \pm 26.2$ & 27\% \\
\midrule
Total & $222$ & 100\% \\
\bottomrule
\end{tabular}
\end{table}